\pdfoutput=1
\documentclass[11pt]{article}

\usepackage[final]{acl}

\usepackage{times}
\usepackage{latexsym}

\usepackage[T1]{fontenc}
\usepackage[utf8]{inputenc}

\usepackage{microtype}

\usepackage{inconsolata}

\usepackage{amsmath}
\usepackage{amsfonts}
\usepackage{graphicx}
\usepackage{amssymb}
\usepackage{booktabs} 
\usepackage{musicography}
\usepackage{multirow}
\usepackage[inline]{enumitem} 
\usepackage{graphicx}
\usepackage{subfig}
\usepackage{xcolor}
\definecolor{LightSteelBlue1}{RGB}{202,225,255}
\definecolor{LightPink}{RGB}{245,191,210}
\definecolor{Moccasin}{RGB}{255, 228, 181}
\usepackage{tikz}
\usepackage{pifont}
\usepackage{xspace}
\usepackage{bm}
\usepackage{tablefootnote}

\usepackage{multirow}

\definecolor{LightSteelBlue1}{RGB}{202,225,255}

\def\method{\textsc{GNNavi}\xspace}
\def\gcn{\textsc{GNNavi-GCN}\xspace}
\def\sage{\textsc{GNNavi-SAGE}\xspace}

\title{\method: Navigating the Information Flow in Large Language Models by Graph Neural Network}
\author{Shuzhou Yuan$^{1,2}$,~Ercong Nie$^{3,4}$,~Michael Färber$^{1,2}$,~Helmut Schmid$^{3}$,~Hinrich Sch\"utze$^{3,4}$ \\
$^{1}$Center for Scalable Data Analytics and 
Artificial Intelligence (ScaDS.AI), Germany \\
$^{2}$TU Dresden, Germany \\
$^{3}$Center for Information and Language Processing (CIS), LMU Munich, Germany \\
$^{4}$ Munich Center for Machine Learning (MCML), Germany \\
\texttt{shuzhou.yuan@tu-dresden.de,~nie@cis.lmu.de}}

\begin{document}
\maketitle
\begin{abstract}

Large Language Models (LLMs) exhibit strong In-Context Learning (ICL) capabilities when prompts with demonstrations are used.
However, fine-tuning still remains crucial to further enhance their adaptability. 
Prompt-based fine-tuning proves to be an effective fine-tuning method in low-data scenarios, 
but high demands on computing resources limit its practicality. 
We address this issue by introducing a prompt-based \textit{parameter-efficient fine-tuning (PEFT)} approach.
\textbf{\method} leverages insights into ICL's information flow dynamics, which indicates that label words act in prompts as anchors for information propagation.
\method employs a \textit{Graph Neural Network (GNN)} layer to precisely guide the aggregation and distribution of information flow during the processing of prompts by hardwiring the desired information flow into the GNN.
Our experiments on text classification tasks with GPT-2 and Llama2 show \method surpasses standard prompt-based fine-tuning methods in few-shot settings by updating just 0.2\% to 0.5\% of parameters. 
We compare \method with prevalent PEFT approaches, such as prefix tuning, LoRA and Adapter in terms of performance and efficiency. Our analysis reveals that \method enhances information flow and ensures a clear aggregation process.
\footnote{Our code is available at \url{https://github.com/ShuzhouYuan/GNNavi}.}
\end{abstract}

\section{Introduction}

\begin{figure}[ht]
    \centering
    \includegraphics[width=1.005\linewidth]{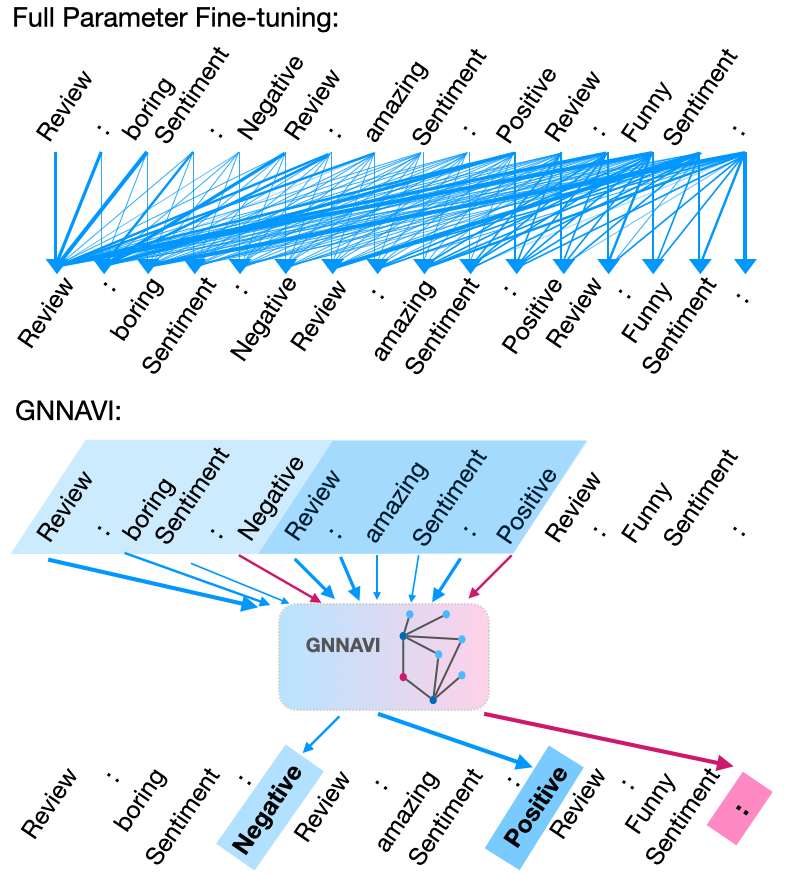}
    \caption{Visualization of Full Parameter Fine-tuning (FPFT) and \method from the perspective of information flow (top words to bottom words). Without \method, tokens interact with every preceding word in FPFT, leading to confusion in information flow. Conversely, in \method, label words aggregate information from preceding words (\colorbox{LightSteelBlue1}{blue path}), and the final token aggregates information from the label words (\colorbox{LightPink}{pink path}), resulting in a clearer information aggregation process.}
    \label{intro_figure}
\end{figure}

% Large Language Models (LLMs) has witnessed exponential growth of model sizes. For instance, GPT2-XL contains 1.5 billion parameters \citep{radford2019language}, and Llama2 comprises 70 billion parameters \citep{touvron2023llama}. While the huge model size has brought state-of-the-art performance on Natural Language Processing (NLP) benchmarks, the required resources for training a LLM are becoming increasingly expensive. This renders standard fine-tuning challenging, as all LLM parameters need to be updated for specific NLP tasks \citep{devlin-etal-2019-bert}.

Large language models (LLMs) show remarkable In-Context-Learning (ICL) capabilities by learning from prompts with demonstrations~\citep{wan-etal-2023-gpt, sun-etal-2023-text, patel-etal-2023-magnifico, mekala-etal-2023-zerotop, ko-etal-2023-large}, 
% alongside notable emergent capabilities~\citep{wei2022emergent, schaeffer2023are}, 
with the exponential growth in model sizes. 
However, fine-tuning LLMs still remains essential for further enhancing their adaptability~\citep{zhang2023llama}. Prompt-based fine-tuning~\citep{schick-schutze-2021-exploiting, ma-etal-2024-topro},
adopting objectives that simulate the language modeling process, emerges as a viable technique, particularly in low-data settings~\citep{gao-etal-2021-making}. Yet, the substantial computational demands of Full-Parameter Fine-Tuning (FPFT), which
updates billions of parameters, pose a practical challenge. 
In fact, optimizing a relatively small subset of an LLM's parameters can significantly improve its performance~\citep{ding2023parameter}, paving the way for Parameter-Efficient Fine-Tuning (PEFT) methods. These methods include Adapter~\citep{pmlr-v97-houlsby19a}, Prompt-Tuning~\citep{lester-etal-2021-power}, Prefix Tuning~\citep{li-liang-2021-prefix}, and LoRA~\citep{hu2022lora}. They offer alternatives to FPFT but are often not tailored
to the prompt-based fine-tuning of LLMs.

Recent advances in understanding the ICL mechanism offer a new avenue for PEFT of LLMs. ICL's success in leveraging few-shot demonstrations and prompts~\citep{brown2020language} has motivated the adoption of prompt-based fine-tuning for moderately sized language models in a few-shot learning manner~\citep{ma-etal-2023-prompt, schick-schutze-2021-just}. 
% In the era of LLMs, instruction tuning has become the \textit{de facto} alignment method to enhance the controllability and performance of LLMs~\citep{zhang2023instruction}. 
Recognizing the specific features of fine-tuning LLMs within the framework of ICL, we propose \textbf{\method}, a novel PEFT method designed expressly for prompt-based learning. Our method draws inspiration from recent insights into the underlying process of ICL from an information flow perspective, particularly the role of label words in the prompt~\citep{wang-etal-2023-label}. 
Label words act as anchors with two functions: aggregating information from context words and directing this information to the last token for accurate predictions. \method incorporates this understanding through the integration of a Graph Neural Network (GNN) layer \citep{kipf2017semisupervised,hamilton2017inductive} into LLMs, optimizing the prompt-based fine-tuning process by navigating the information flow within prompts, as visualized in Figure \ref{intro_figure}.
Following the paths of information flow, we insert a GNN layer into the deep layers\footnote{We use ``deep layers'' to refer to the last few layers of the LLM. For instance, in GPT2-XL, there are 48 layers, with the last 12 layers considered as deep layers in our work.} of the LLM. We treat the input text as a graph, where each token serves as a node, and connect these nodes according to the paths of information flow. 
% The node features are initialized with the token representation from the last layer of the LLM directly preceding the GNN layer. 
The GNN layer aims to guide the information flow by aggregating information from neighbouring nodes.

As a PEFT method, \method adopts a lightweight fine-tuning strategy, updating only the parameters of the GNN layer. Experimenting with few-shot training examples on GPT2-XL \citep{radford2019language} and Llama2 \citep{touvron2023llama}, \method achieves remarkable results with just 0.2\% of the trainable parameters of the full model, consistently outperforming FPFT and other PEFT methods across various classification tasks. Additionally, we analyze the attention interaction between tokens and find that \method demonstrates a more stable and clear information aggregation process compared to FPFT.

In summary, our contributions are:
\begin{enumerate*}[label={\textbf{\roman{*})}}]
    \item We propose a novel PEFT method, \method, inspired by the information flow perspective of LLMs. \method effectively navigates the information aggregation process in LLMs.
	
    \item We apply \method to text classification tasks with few-shot training examples, outperforming baselines while updating only 0.2\% to 0.5\% of parameters.

    \item Our work sheds light on the application of GNNs in NLP and provides novel insights for future research. To the best of our knowledge, we are the first to utilize GNNs to enhance the performance of LLMs from the information flow perspective. 
\end{enumerate*}

\section{Related Work}
	\begin{figure*}[htbp]
    \centering
    \includegraphics[width=1\textwidth]{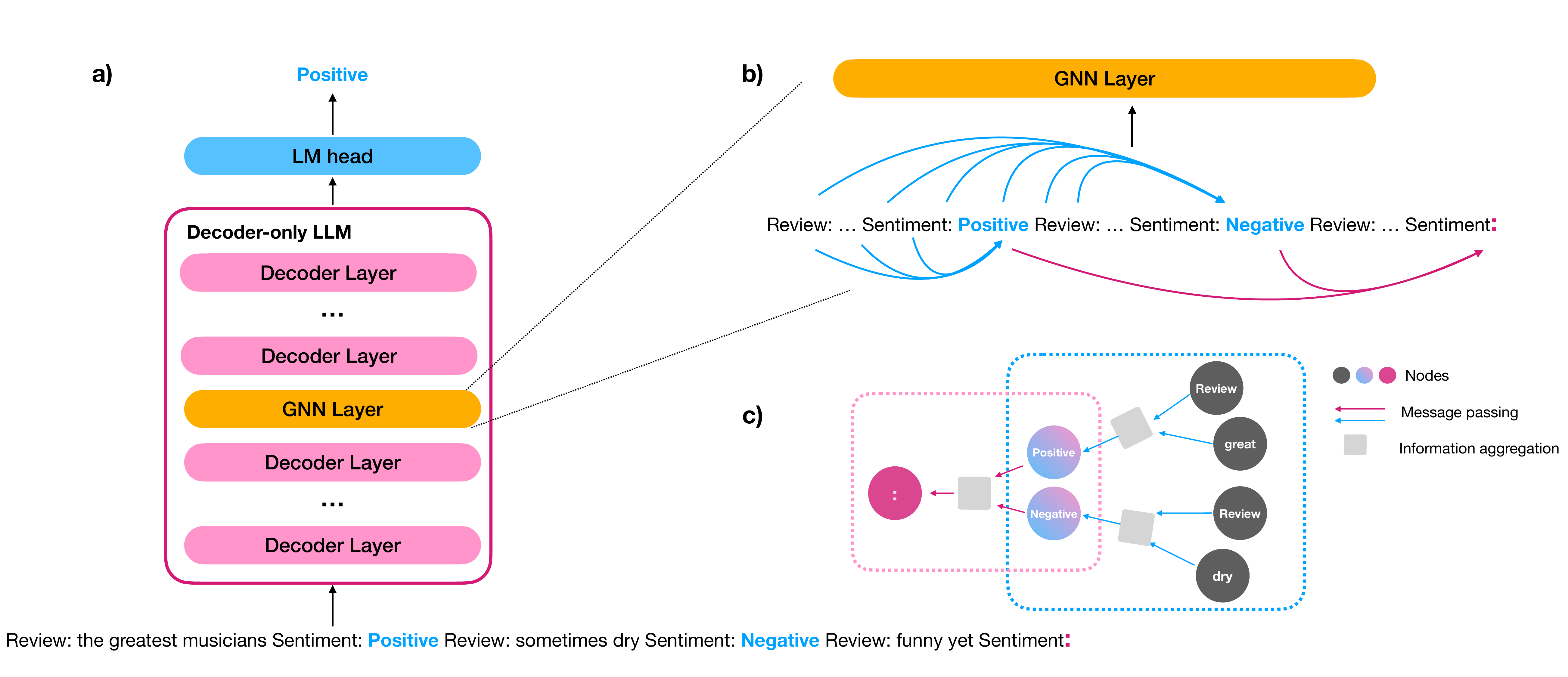}
    \caption{Visualization of \method with an example of sentiment analysis, where label words and the last token are highlighted in blue and pink, respectively. a) The GNN layer is integrated into a decoder-only LLM. The LLM processes a prompt containing demonstrations and generates the next token as the prediction. b) The input text is transformed into a graph, with tokens as nodes and information flow paths as edges. c) Visualizing the working mechanism of the GNN: Node representations are updated by aggregating information from neighboring nodes. To maintain simplicity, not all nodes are listed.}
    \label{method_overview}
\end{figure*}

\paragraph{Prompt-Based Learning}
GPT-3~\cite{brown2020language} has sparked interest in prompt-based learning methods, and particularly in the ICL paradigm. This surge in attention has fostered a multifaceted exploration into the factors influencing ICL performance, including input perturbation~\citep{yoo-etal-2022-ground, min-etal-2022-rethinking}, selection of demonstration~\citep{liu-etal-2022-makes, nie-etal-2023-cross}, and calibration techniques~\citep{zhao2021calibrate, nie-etal-2023-unleashing}. Concurrently, there has been a deep dive into understanding the underlying mechanism of ICL, employing diverse theoretical frameworks such as gradient descent~\citep{dai-etal-2023-gpt}, Bayesian inference~\citep{xie2022an} and information flow~\citep{wang-etal-2023-label}. Following the route of ICL, prompt-based fine-tuning has emerged as an effective strategy in scenarios with limited data~\citep{gao-etal-2021-making, schick-schutze-2021-exploiting, schick-schutze-2021-just}. We leverage insights from these investigations into the ICL mechanism and propose a tailored PEFT method for LLMs.

\paragraph{Parameter-Efficient Fine-Tuning (PEFT)}
PEFT focuses on enhancing language model performance on downstream tasks by optimizing a small number of parameters, instead of fine-tuning all parameters~\citep{ding2023parameter}.
Various PEFT strategies have been explored. Addition-based methods only train modules or parameters added to the model, such as Adapter~\citep{pmlr-v97-houlsby19a}, Prompt tuning~\citep{lester-etal-2021-power}, and Prefix tuning~\citep{li-liang-2021-prefix}. 
Specification-based methods selectively fine-tune specific parameters in the original model while keeping the remainder frozen, such as BitFiT~\citep{ben-zaken-etal-2022-bitfit}. 
Reparameterization-based methods transform existing parameters into a more parameter-efficient form, such as LoRA~\citep{hu2022lora}.
Recent advancements in PEFT research have increasingly prioritized memory efficiency, aiming to enable the training of LLMs with minimal computational resources, such as MeZO~\citep{malladi2023finetuning} and HiFT~\citep{liu2024hift}.
Our proposed PEFT method is designed specifically for LLMs and draws upon the intricacies of how LLMs process and learn from prompts.

\paragraph{GNN for NLP} GNNs are predominantly utilized in NLP tasks involving structural input, such as graph-to-text generation \citep{gardent-etal-2017-webnlg} and graph-enhanced question answering \citep{zhang2022greaselm}. Previous approaches employ GNNs to encode complex graph and node representations. For instance, \citet{koncel-kedziorski-etal-2019-text} introduced Graph Transformer, which extends graph attention networks \citep{velickovic2018graph} for encoding scientific graph inputs, while \citet{li-etal-2021-shot-knowledge} utilize GNNs to encode knowledge graphs and align them with text embeddings from pretrained language models. Additionally, GNNs serve as auxiliary tools for pretrained language models to encode complex structural information for AMR-to-text generation \citep{ribeiro-etal-2021-structural}. Unlike prior work, we leverage GNNs for information aggregation based on the perspective of information flow.

\section{Method}
% In this section, we elaborate on the details of our approach. We begin by detailing the architecture of \method, followed by the task formulation.

\subsection{Architecture of \method}

\paragraph{Intuition} \citet{wang-etal-2023-label} demonstrated that the working mechanism of LLM follows specific paths of information flow. The label words in the input prompt serve two roles for the final predictions: acting as information aggregators by gathering information from their preceding words and propagating the aggregated information to the last token position where the prediction is generated. Building upon their insights, we posit that navigating the flow of information aggregation can enhance both efficiency and effectiveness of LLMs. Leveraging the GNN's proficiency in information aggregation at the graph level, we explore LLMs from a graph theory perspective and utilize GNN as a tool to guide the information flow.

\paragraph{Working Mechanism}  
% We integrate a GNN layer with the LLM by adding it subsequent to the $l$-th decoder layer. Unlike previous work that incorporated the Adapter module into language models by adding it to every decoder layer \citep{ribeiro-etal-2021-structural,wang-etal-2021-k}, we only add one GNN layer after one specific decoder layer in LLM. Our approach further reduces training costs.
% as shown in Figure \ref{model_integration}. Therefore, the token representations obtained from its last decoder layer is fed into the GNN layer directly. 
% \begin{figure}[htbp]
%     \centering
%     \includegraphics[width=0.3\textwidth]{images/model_integretation.png}
%     \caption{Integration of GNN layer with decoder layer. The GNN layer is added after the add and normalization sub-layer of a decoder layer.}
%     \label{model_integration}
% \end{figure}
We illustrate the working mechanism of \method in Figure \ref{method_overview}. For example, in a sentiment analysis task, the prompt comprises one demonstration from each class and the text to be classified. An LLM processes this prompt layer by layer. The GNN layer is inserted after the $l$-th decoder layer of the LLM\footnote{In our preliminary experiments, \method performs optimally when the GNN layer is inserted in the last quarter of the layers in LLM. Thus, we add the GNN layer after the 42nd layer of GPT2-XL and after the 28th layer of Llama2-7b in our experiments. A detailed analysis is conducted in \S\ref{ana_position}.}. Receiving the token representations from the $l$-th layer, the GNN layer learns node representations by aggregating information from neighboring nodes. 
Subsequently, the node representations are propagated to the next layer in LLM as hidden states. 
The nodes are connected following the paths of information flow. As depicted in Figure \ref{method_overview}(b), the label words `\textit{Positive}' and `\textit{Negative}' aggregate information from their preceding tokens
and pass the information to the last token `\textit{:}' of the prompt. In case the label word is tokenized into subtokens, we use the first subtoken to serve as the label word, following previous work \citep{pmlr-v139-zhao21c,wang-etal-2023-label}. We freeze the pretrained parameters of the LLM during training and update only the parameters in the GNN layer.

\paragraph{Graph Neural Network} The graph neural network aggregates information from neighboring nodes to model graph and node representations by message passing. To formulate an NLP task on a graph level, we consider the input text as a graph.
We define a directed graph $\mathcal{G}$ as a triple $(\mathcal{V}, \mathcal{E}, \mathcal{R})$ with a set of nodes $\mathcal{V} = \{v_1, \ldots, v_n\}$ (one node for each token), a set of relation types $\mathcal{R}$\footnote{In our work, we only consider one relation type: the directed edge from node $v$ to node $v^{\prime}$.},
%HS? How is the relation set defined?
and a set of edges \( \mathcal{E} \) of the form \( (v, r, v^{\prime}) \) with $v, v' \in \mathcal{V}$, and $r \in \mathcal{R}$.
Each node $v_i$ is associated with a feature vector $x_i$, which is the token representation of the $i$-th token in the $l$-th layer. In Figure \ref{method_overview}, for instance, the first token `\textit{Review}' is connected with the label token `\textit{Positive}'. This edge is represented by the triple \( (\text{\textit{Review}}, aggregate, \text{\textit{Positive}}) \), where $aggregate$ denotes an edge directed towards a label node.

The node representations in GNN layer are updated by aggregating the information from neighboring nodes. The aggregation algorithms vary across different GNN architectures. For example, the learning process of Graph Convolutional Network (GCN) \citep{kipf2017semisupervised} is formulatd as:
\begin{equation}\footnotesize
  h_v = \sigma\left(W\sum_{v^{\prime} \in N(v)} \frac{h_{v^{\prime}}^{(l)}}{\lvert N(v)\rvert}\right)
\end{equation} 
where $h_v$ denotes the updated node representation of $v$, $h_{v^{\prime}}^{(l)}$ denotes the token representation of its neighbouring nodes from $l$-th decoder layer, $\sigma$ is the activation function, $W$ is the trainable parameter of GNN, $N(v)$ includes all the neighbouring nodes of $v$. 

We also include another GNN architecture, GraphSAGE \citep{hamilton2017inductive}, in our studies, which involves a more complex learning process:
\begin{equation}\footnotesize
  h_v = \sigma\left(W\left(h_v^{(l)}\oplus\text{AGG}(\{h_{v^{\prime}}^{(l)}, \forall v^{\prime} \in N(v)\}) \right)\right)
\end{equation} 
The concatenation function $\oplus$ concatenates aggregated information with the node current representation, and the aggregation function AGG compiles message passing from neighboring nodes using techniques such as mean, pool and LSTM.\footnote{We apply mean aggregation to GraphSAGE in this work.}
% \begin{equation}
%   \text{AGG} = \sum_{\forall v^{\prime} \in N(v)} \frac{h_{v^{\prime}}^{(l-1)}}{\lvert N(v)\rvert}
% \end{equation} 
We visualize the information aggregation process of GNN in Figure \ref{method_overview}(c).

\subsection{Task Formulation}
In our work, we implement prompt-based fine-tuning for text classification tasks.
Our goal is to predict the correct class given a few examples. We reformulate the task as a language modeling problem. Let $M$ be a language model with vocabulary $V$, and let $\mathcal{L}$ be a set of label words. The training set $\mathcal{T}$ consists of pairs $(s, l)$, where
$s$ is a sequence of tokens from the vocabulary $V$ and $l$ is a label word from the set $\mathcal{L}$. In a sentiment analysis task, for instance, we define a pattern $\mathcal{P}(s,l)$ which associates a text $s=$`Nice performance' and a label word $l=$`Positive' as follows:
$$
\colorbox{LightSteelBlue1}{\text{Review: \underline{Nice performance.} Sentiment: \underline{Positive}}}
$$
For a $k$-class classification task, we sample one demonstration per class from the training set $\mathcal{T}$, and concatenate them with the text $s$ to be classified to form the prompt $X(s)$:
\begin{equation}
\label{equa_prompt}
X(s) = \mathcal{P}(s_1,l_1)\oplus \ldots \oplus \mathcal{P}(s_k,l_k)\oplus \mathcal{P}(s,\varepsilon)
\end{equation}
$\oplus$ denotes the concatenation of the input demonstrations and $\varepsilon$ is the empty string. A more intuitive example is shown in Figure \ref{method_overview}.
The language model reads %Taking
the prompt $X(s)$ and predicts the next token $l$, 
which is the label assigned to $s$. $M$ is initialized with pretrained parameters $\phi$, and fine-tuned by minimizing the cross-entropy loss: 
\begin{equation}
  \ell = -\sum_{(s,l) \in \mathcal{T}} \log p_\phi(X(s), l)
\end{equation} 
$p_\phi(.,.)$ returns the probability which $M$ assigns to the correct label $l$. In our work, we randomly select one demonstration per class to form the prompt and remove them from $\mathcal{T}$. The training examples are then sampled from the remaining samples in $\mathcal{T}$.

\section{Experiments}

\subsection{Datasets}

We implement text classification tasks using five commonly used datasets from different domains, including
\textbf{SST-2:} Stanford Sentiment Treebank Binary for sentiment analysis \citep{socher-etal-2013-recursive};
\textbf{EmoC:} EmoContext for 4-label emotion classification \citep{chatterjee-etal-2019-semeval};
\textbf{TREC:} Text REtrieval Conference Question Classification (TREC) for question type classification containing 6 types \citep{li-roth-2002-learning,hovy-etal-2001-toward};
\textbf{Amazon:} binary classification for Amazon reviews \citep{mcauley2013hidden};
\textbf{AGNews:} AG’s news topic classification dataset for topic classification with 4 labels \citep{NIPS2015_250cf8b5}.

\subsection{Experimental Setting\label{exp_setting}}

The prompt is designed following the template in Equation \ref{equa_prompt}. We take one demonstration per class to form the prompt\footnote{The templates of prompts are presented in Appendix \ref{template}.} and append the sample to be predicted at the end of the prompt. Following a few-shot learning setting, we experiment with different numbers of training samples, namely 5, 10, 20, 50, 100, and 200 samples per class. The training samples are randomly selected from the original training set. Another 1000 samples from the original training set are sampled as the validation set, and 1000 samples from the original test set are used for evaluation.\footnote{The original test set of SST-2 contains less than 1000 samples, so we keep the original test set for model evaluation.}
The accuracy on the validation set is employed to identify the best-performing model, which is subsequently evaluated on the test set. We report the average accuracy over five random seeds. The hyperparameters can be found in Appendix \ref{hyp}.

\subsection{Models}
As \method is built on the base of decoder-only LLMs, we select two large language models, both with over 1 billion parameters, and equip them with \method. Specifically, we choose GPT2-XL with 1.6 billion parameters \citep{radford2019language} and Llama2 with 7 billion parameters \citep{touvron2023llama}. For the GNN layer, we opt for GCN and GraphSAGE, denoted as \textbf{\gcn} and \textbf{\sage} in the experiments. To integrate \method with GPT2-XL and Llama2, we modify their source codes from Huggingface \citep{wolf2019huggingface} and utilize GNN models provided by PyTorch Geometric \citep{Fey/Lenssen/2019}.

\subsection{Baselines}
% We adopt the following baselines for the experiments:

\textbf{ICL one-shot per class:} In-context learning (ICL) follows the scenario where the LLM is initialized with pre-trained parameters and instructed by demonstrations to perform text classification tasks. None of the model parameters are updated. We sample one demonstration per class to form the prompt. The demonstrations used to form the prompt are consistent with those used for other methods under the same random seed.

\textbf{ICL few-shot per class:} To compare with the low-data fine-tuning setting, we implement ICL with 5 additional shots per class as the demonstrations. This setting is comparable to a training set with a size of 5 samples per class. Due to the limited input length of GPT2-XL, AGNews and Amazon are set to 4 additional shots per class.

\textbf{Low-Rank Adaptation (LoRA):} LoRA is a PEFT method that reduces the number of trainable parameters by injecting trainable rank decomposition matrices into each layer of the LLM \citep{hu2022lora}. We implement LoRA using the Python library PEFT \citep{peft}.

\textbf{Prefix-tuning (Prefix):} Prefix-tuning utilizes a soft-prompt strategy, incorporating virtual tokens into the LLM and updating only the parameters of the virtual tokens \citep{li-liang-2021-prefix}. We implement prefix-tuning using the PEFT library \citep{peft}. The number of virtual tokens\footnote{The number of virtual tokens can be found in Appendix \ref{hyp}.} is set to maintain a comparable size of trainable parameters as for \method.

\textbf{Adapter:} We insert a standard adapter module after the feed-forward sub-layer of each layer in the LLM \citep{pmlr-v97-houlsby19a}. The adapter module is added using AdapterHub \citep{pfeiffer2020AdapterHub,poth-etal-2023-adapters}.

\textbf{Full Parameter Fine-tuning (FPFT):} Full parameter fine-tuning is implemented as a strong baseline, where all the model parameters are updated during the training process.

\section{Results}
\begin{table*}[ht]
\centering
\scalebox{0.58}{
\begin{tabular}{crcccccc|rcccccc} 
\toprule
\textbf{Method} & \textbf{\#Param} & \textbf{SST-2} & \textbf{EmoC}  & \textbf{TREC}  & \textbf{Amazon} & \textbf{AGNews} & \textbf{Average} & \textbf{\textbf{\#Param}} & \textbf{SST-2} & \textbf{EmoC}  & \textbf{TREC}  & \textbf{Amazon} & \textbf{AGNews} & \textbf{Average}  \\ 
\midrule
                & \multicolumn{7}{c|}{\textbf{GPT2-XL}}                                                                                      & \multicolumn{7}{c}{\textbf{Llama2}}                                                                                                  \\ 
\midrule
                & \multicolumn{14}{c}{$k=0$}                                                                                                                                                                                                                                          \\ 
\midrule
ICL             & -                & 55.44          & 6.48           & 54.68          & 53.32           & 72.12           & 48.41            & -                         & 67.55          & 9.60           & 70.36          & 94.98           & 84.14           & 65.33             \\ 
\midrule
                & \multicolumn{14}{c}{$k=5$}                                                                                              \\ 
\midrule
ICL             & -                & 63.17          & 6.30           & 57.68          & 53.67           & 50.43           & 46.25            & -                         & 86.93          & 20.18          & 45.72          & 92.30           & 80.16           & 65.06             \\ 
\midrule
LoRA            & 2.5M             & 91.98          & 50.60          & 75.20          & 88.80           & \textbf{85.20}  & 78.36            & 4.2M                      & \textbf{95.42} & 64.20          & \textbf{88.40} & 91.80           & 86.60           & 85.28             \\
Prefix          & 6.1M             & 59.13          & 73.46          & 32.92          & 60.00           & 75.40           & 60.18            & 39.3M                     & 50.96          & 58.56          & 21.36          & 49.36           & 25.78           & 41.20             \\
Adapter         & 15.4M            & 79.82          & 76.00          & \textbf{79.60} & \textbf{91.45}  & 81.25           & 81.62            & 198M                      & 50.92          & \textbf{84.05} & 18.80          & 49.45           & 24.80           & 45.60             \\
FPFT            & 1.6B             & 62.13          & 61.30          & 65.28          & 73.00           & 80.82           & 68.51            & 6.7B                      & 94.63          & 61.92          & 81.72          & \textbf{95.86}  & \textbf{87.58}  & 84.34             \\ 
\midrule
  \gcn              & 2.6M             & \textbf{84.31} & 75.48          & 76.72          & 90.90           & 83.16           & \textbf{82.11}   & 16.8M                     & 94.56          & 78.30          & 83.2           & 94.00           & 86.25           & 86.63             \\
    \sage            & 5.1M             & 81.95          & \textbf{78.70} & 77.92          & 88.66           & 82.88           & 82.02            & 33.6M                     & 92.91          & 80.12          & 80.80          & 95.66           & 86.06           & \textbf{87.11}    \\ 
\midrule
               & \multicolumn{14}{c}{$k=200$}                                                                                                   \\ 
\midrule
LoRA            & 2.5M             & \textbf{90.83} & 80.80          & 90.80          & 82.00           & 86.20           & 86.13            & 4.2M                      & 91.29          & \textbf{86.80} & 93.60          & 95.80           & 90.40           & 91.32             \\
Prefix          & 6.1M             & 50.92          & 80.18          & 69.80          & 59.80           & 79.08           & 67.96            & 39.3M                     & 48.35          & 81.72          & 45.68          & 52.28           & 27.54           & 51.11             \\
Adapter         & 15.4M            & 88.65          & 80.70          & \textbf{96.60} & 92.30           & \textbf{89.80}  & 89.61            & 198M                      & 50.92          & 85.05          & 88.20          & 49.45           & 81.50           & 67.57             \\
FPFT            & 1.6B             & 68.97          & 73.70          & 80.16          & 74.82           & 85.34           & 76.60            & 6.7B                      & \textbf{95.64} & 79.90          & \textbf{96.76} & 96.12           & \textbf{91.44}  & 91.97             \\ 
\midrule
 \gcn               & 2.6M             & 90.67          & 78.82          & 91.88          & 92.94           & 89.20           & 88.70            & 16.8M                     & 95.36          & 82.85          & 95.50          & \textbf{96.45}  & 91.05           & \textbf{92.24}    \\
  \sage              & 5.1M             & 90.46          & \textbf{82.68} & 92.32          & \textbf{93.44}  & 89.28           & \textbf{89.64}   & 33.6M                     & 95.30          & 81.94          & 94.76          & 95.96           & 90.68           & 91.73             \\
\bottomrule
\end{tabular}}
\caption{Results of different training methods (accuracy). $k$ denotes the number of training examples per class,  \#Param denotes the number of trainable parameters. The best scores are highlighted with \textbf{bold}.}
\label{results}
\end{table*}
\begin{figure*}[hbt]
    \centering
    \includegraphics[width=1\textwidth]{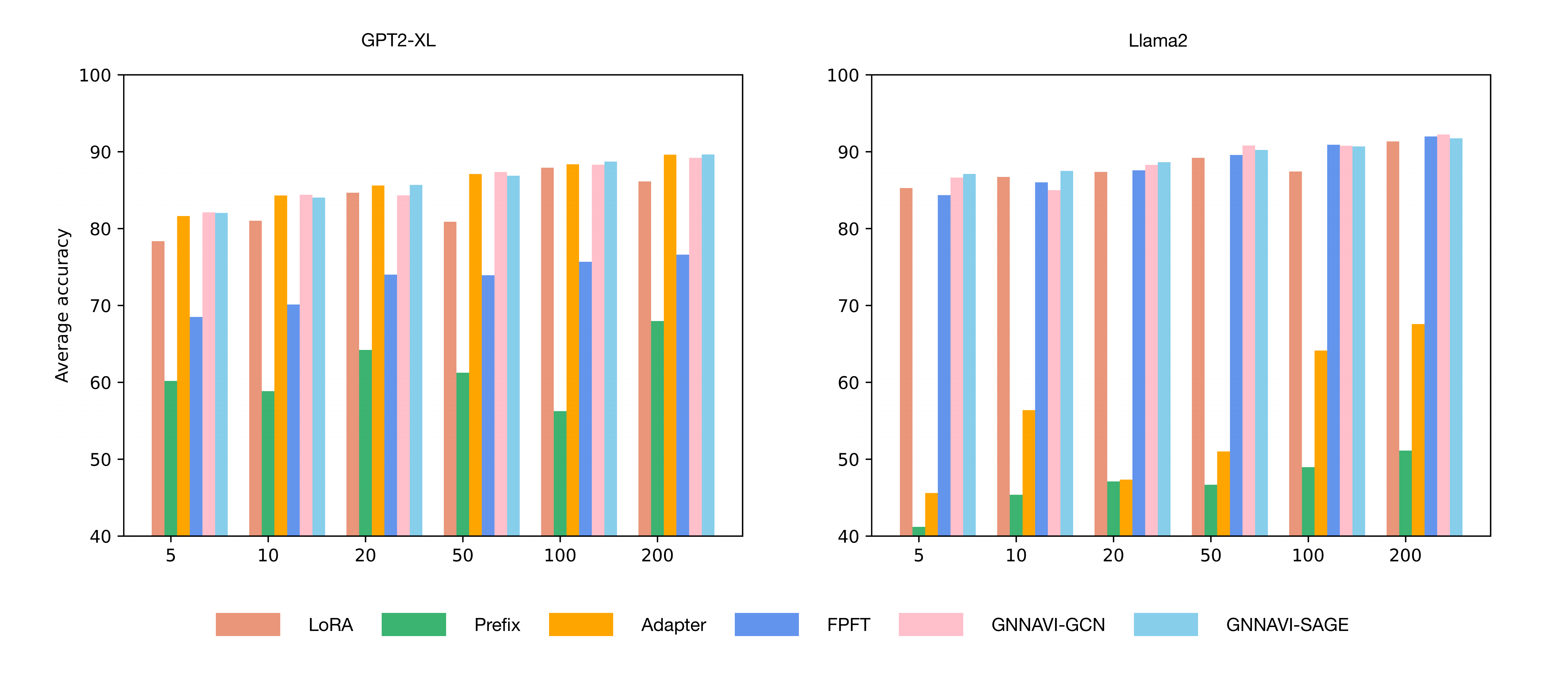}
    \caption{Results of average accuracy with different number of training examples. The x-axis denotes the number of training examples per class.}
    \label{results_bar}
\end{figure*}

We report the results with 5 and 200 training examples in Table \ref{results}, which reflect the performance under the scenarios where only limited training examples are available and sufficient training examples are provided respectively. 
% The results with different training sizes are visualized in Figure \ref{results_bar}. 
Full results are presented in Appendix \ref{full_results_app}.

\subsection{Overall Performance}

Observing the results of GPT2-XL, \method remarkably rivals ICL, FPFT, and other parameter-efficient baselines. Under the low-data setting of 5 training examples, both \gcn and \sage outperform FPFT by over 13\%, achieving higher accuracy than other PEFT methods by 0.4\% to 21\%. Increasing the number of training examples to 200, the average performance of \method improves to 89.64\% and outperforms other baselines.

Similar to GPT2-XL, \method achieves the best performance with Llama2 among all the baselines. With only 5 training examples, \sage achieves 2.77\% higher average accuracy than FPFT. Comparing with other PEFT methods, \method shows higher average accuracy from 1.8\% to 35\%. And with 200 training examples, \gcn achieves 92.24\% average accuracy, outperforming FPFT, Prefix-tuning, Adapter, and LoRA.

\subsection{Efficiency Analysis}
\begin{table}[htb]
\centering
\scalebox{0.7}{
\begin{tabular}{cccccc} 
\toprule
        & \textbf{SST-2} & \textbf{EmoC} & \textbf{TREC} & \textbf{Amazon} & \textbf{Agnews}  \\ 
\midrule
GPT2-XL & 4.7$\times$            & 6.3$\times$           & 4.1$\times$            & 3.9$\times$              & 3.4$\times$               \\
Llama2  & 4.3$\times$             & 2.4$\times$            & 1.6$\times$            & 1.4$\times$              & 1.2$\times$               \\
\bottomrule
\end{tabular}}
\caption{The ratio by which the training process is accelerated for one training epoch for \gcn compared to FPFT.}
\label{speed_up}
\end{table}

\method significantly reduces the number of trainable parameters
compared to the baselines %
for both GPT2-XL and Llama2.
% Although Prefix, Adapter, and LoRA also reduce the trainable parameters,
\gcn for GPT2-XL
achieves the highest average accuracy with 5 training examples containing
only 2.5 million trainable parameters,
%and achieves the highest accuracy,
which is 615 times smaller than FPFT, six times smaller than Adapter, twice smaller than Prefix, and similar to LoRA. As for Llama2, \method saves over 6.6 billion trainable parameters compared to FPFT and achieves better results. \gcn also updates fewer parameters than Prefix and Adapter. Although LoRA contains fewer trainable parameters than \gcn in Llama2, the performance of LoRA cannot compete with \gcn and \sage.
Table \ref{speed_up} shows that by saving a significant amount of training parameters, \gcn speeds up the training process by a factor of up to 6 compared to FPFT. 
% As shown in Table \ref{speed_up}, by saving a significant amount of training parameters, \gcn is able to speed up the training process up to 6 times compared with FPFT. 

\subsection{Influence of Training Examples \label{train_size}}
	\begin{figure}[htb]
 \centering
		\subfloat[][GPT2-XL]{{\includegraphics[width=6cm]{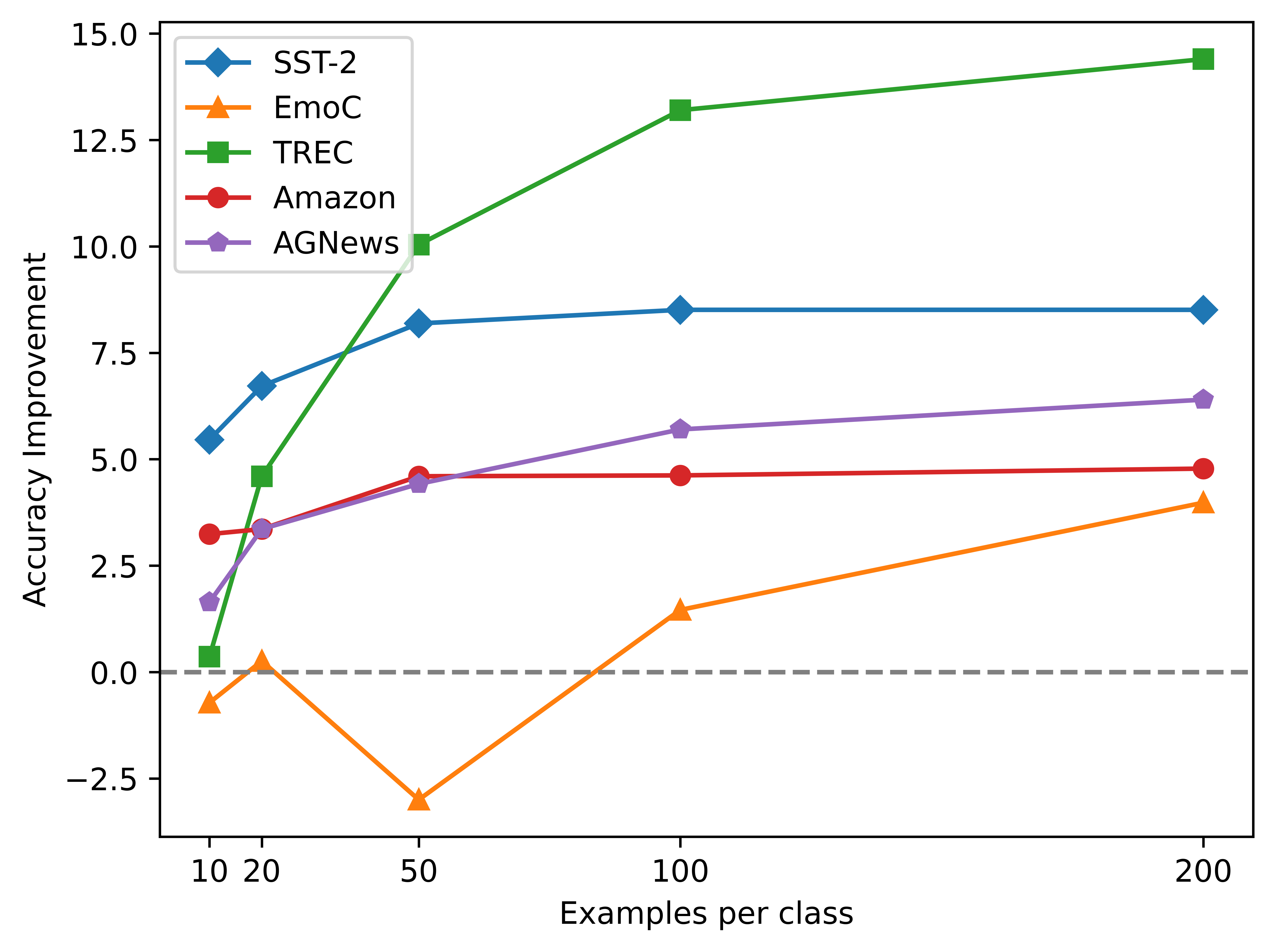} }}%
  
		\subfloat[][Llama2]{{\includegraphics[width=6cm]{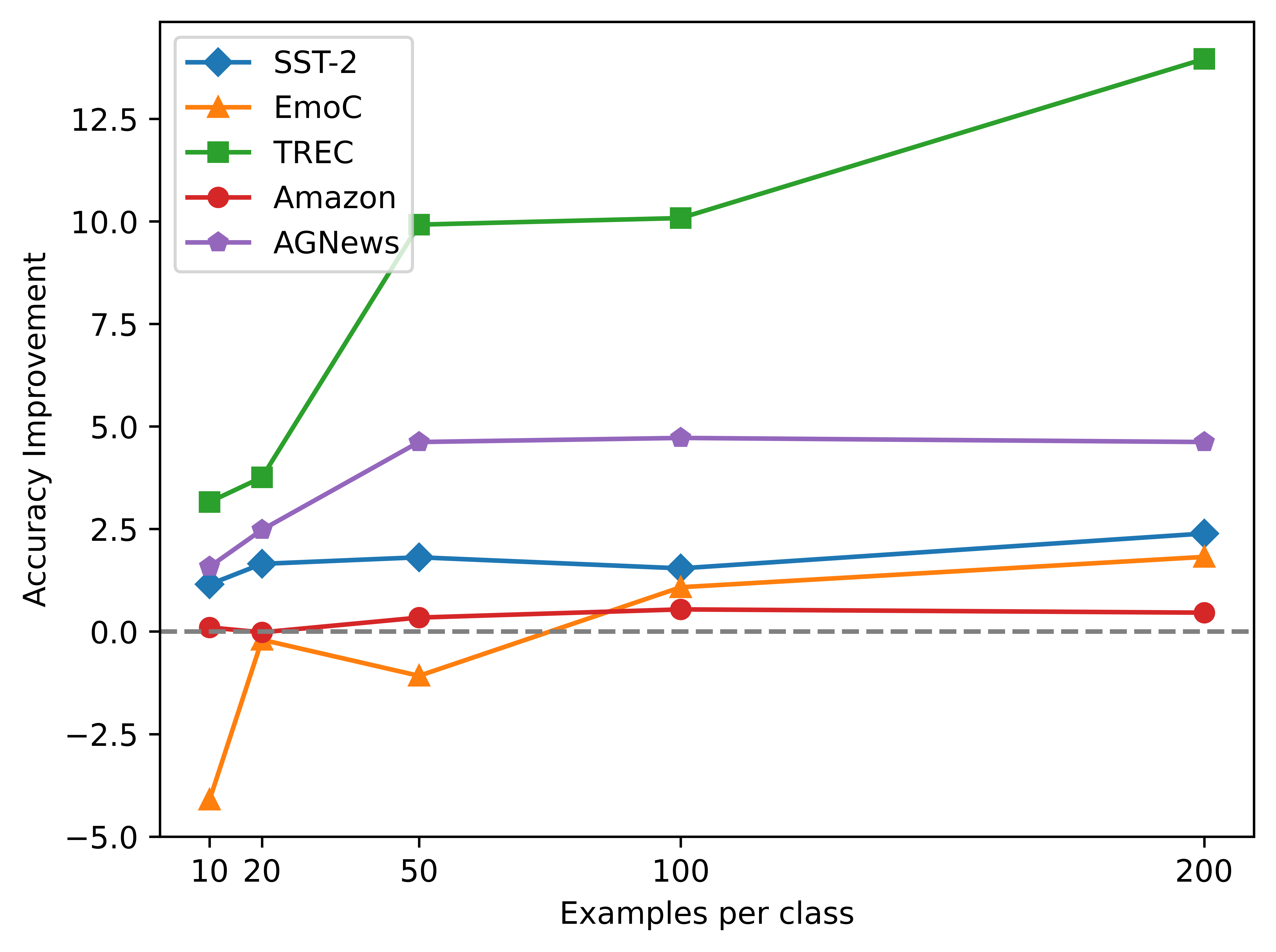} }}%
		\centering\caption{The improvement gained by adding training examples for \sage, compared to using 5 training examples per class.}\label{train_size_figure}
	\end{figure}

Adding more training examples improves the accuracy for most baselines and \method. As depicted in Figure \ref{results_bar}, \method consistently outperforms other methods as the number of training examples increases. While other methods also show improvement with more training examples, the extent of improvement is not as consistent as for \method, particularly for Prefix and Adapter.

Figure \ref{train_size_figure} shows the performance of \method for the different tasks as a function of the number of training examples.
%To analyze the impact of adding training examples for \method, we illustrate the improvements for individual tasks in Figure \ref{train_size_figure}.
We observe that the effect of adding training examples is similar for both GPT2-XL and Llama2. Notably, adding more training examples yields significant improvements, especially in low-data settings (e.g.\ with 10, 20, and 50 training examples) where \method shows a substantial improvement, except for EmoC. However, the significance diminishes when more than 50 training examples are provided, the improvement is not as pronounced here as in low-data settings.

\section{Ablation Study}
In \S\ref{ana_position} of this section, we delve into the influence of the position where the GNN layer is inserted in the LLM. %Additionally,
In \S\ref{remove_edge}, we investigate the effects of removing one of the information flow paths on performance. All of these studies are conducted using \sage with 5 training samples per class under the experimental settings outlined in \S\ref{exp_setting}.

\subsection{Position of GNN Layer\label{ana_position}}
\begin{figure}[ht]
    \centering
    \includegraphics[width=0.45\textwidth]{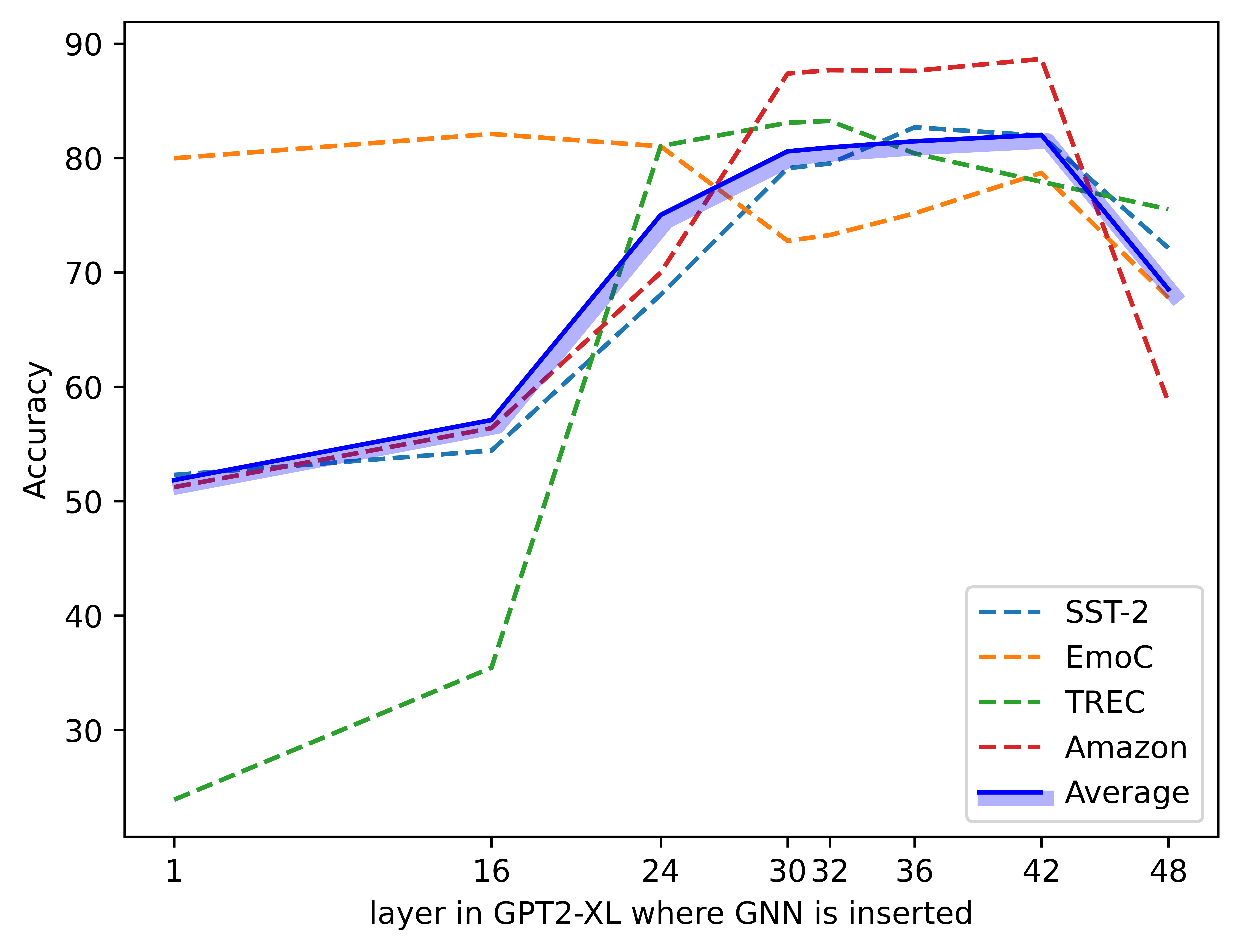}
    \caption{Performance Comparison with GNN inserted at various positions in GPT2-XL.}
    \label{gnn_pos}
\end{figure}
The position where the GNN layer is inserted significantly impacts the model's performance. Figure \ref{gnn_pos} illustrates the performance of \method when the GNN layer is inserted at different locations in GPT2-XL. With the exception of EmoC, all tasks exhibit lower performance when the GNN layer is added in the first 10 layers of GPT2-XL. Performance improves as the GNN is added in deeper layers, reaching peak accuracy around the 44th layer. Subsequently, accuracy declines until the last layer. This trend may stem from the gradual initiation of the information flow process in the early layers of LLM, where the GNN's influence is limited due to insufficient token interaction. Conversely, in the final layers, the information flow process is nearly complete, rendering it too late for the GNN to guide effectively. Despite variations in performance changes across tasks, the average performance suggests that the optimal placement for the GNN layer is between the 38th and 42nd layers for GPT2-XL.

\subsection{Removal of Information Flow\label{remove_edge}}
\begin{figure}[ht]
    \centering
    \includegraphics[width=0.45\textwidth]{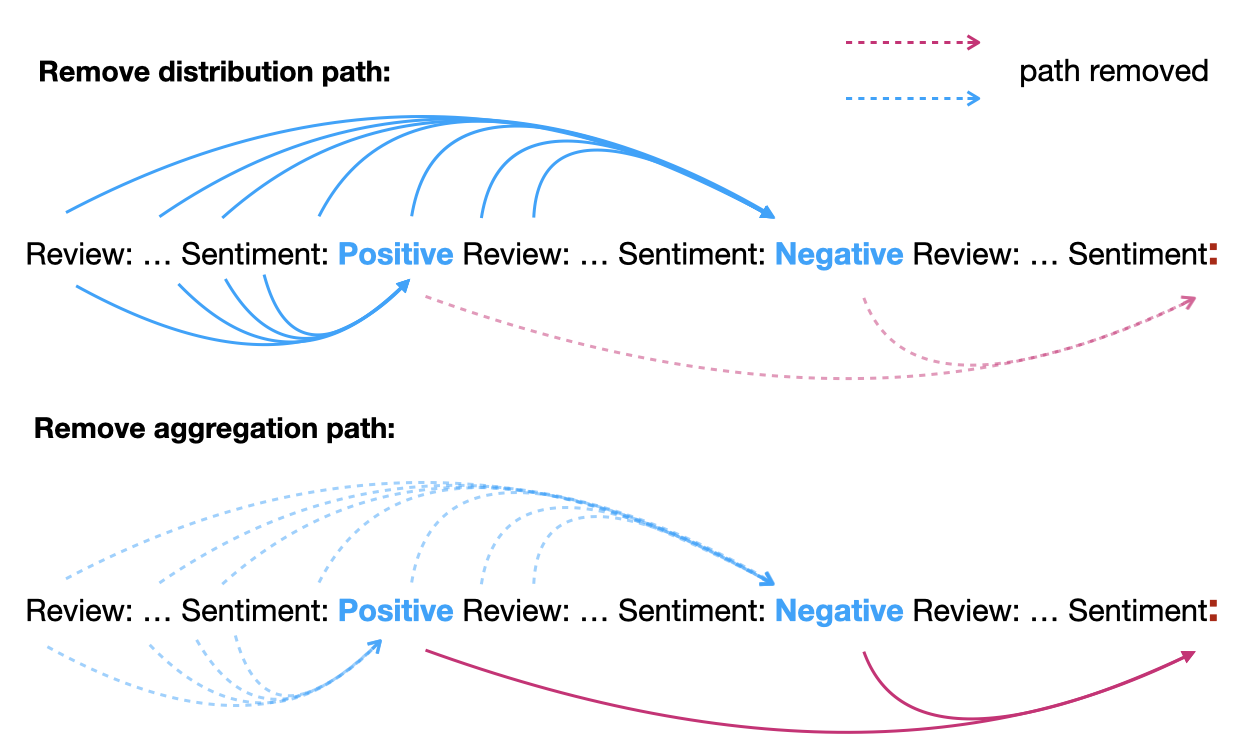}
    \caption{Visualisation of the ablation study on the removal of information flow.}
    \label{ab_path}
\end{figure}

We conduct an ablation study to investigate how removing specific information flow paths affects the results while retaining others. In our approach, we connect the label words to their preceding words to aggregate information and to the last token to distribute the information from the label words. These connections are referred to as the aggregation and distribution paths in the ablation study. As illustrated in Figure \ref{ab_path}, we remove one path and retain another.

\begin{table}[htb]
\centering
\scalebox{0.6}{
\begin{tabular}{ccccccc} 
\toprule
                         & \textbf{SST-2}            & \textbf{EmoC}             & \textbf{TREC}             & \textbf{Amazon}           & \textbf{Agnews}            & \textbf{Average}  \\ 
\midrule
\textbf{\sage} & 81.95 & 78.70 & 77.92 & 88.66 & 82.88 & 82.02             \\
\textbf{-aggregation}            & -0.07                     & -1.10                     & -0.68                     & +0.56                     & -0.08                      & -0.27             \\
\textbf{-distribution}           & +3.07                     & -12.88                    & -2.44                     & +1.64                     & -1.44                      & -2.41             \\
\bottomrule
\end{tabular}}
\caption{Ablation Study: Removal of information flow. The name indicates the removed path.}
\label{ab_path_results}
\end{table}
As shown in Table \ref{ab_path_results}, both the aggregation and distribution paths contribute significantly to the performance. Removing either of them results in a decrease in the average accuracy across the five tasks. Except for the two binary classification tasks SST-2 and Amazon, removing the distribution path causes a greater drop in performance. Based on these results, we conclude that the distribution path plays a more significant role in the information flow process, especially for tasks with more than two labels.

\section{Further Discussion: Information Flow\label{infor_flow}}

While the attention mechanism in LLM offers an information flow perspective for interpreting the model's working mechanism \citep{wang-etal-2023-label}, it treats the input text as a fully connected graph. In contrast, \method
explicitly connects the context tokens to the label tokens for information aggregation and the label tokens to the final token for information distribution. Thereby, the correct information flow is hardwired into the GNN. There is no need to learn it by adjusting the attention weights.
% extracts explicit connectivity between tokens to aggregate information among key tokens, avoiding adjusting the attention weight of every two tokens.
To further investigate the differences in information flow between FPFT and \method, we utilize the saliency technique \citep{journals/corr/SimonyanVZ13} for interpretation. Following the approach of \citet{wang-etal-2023-label}, we compute the saliency score for each element of the attention matrix using a Taylor expansion \citep{NEURIPS2019_2c601ad9}:
\begin{equation}
    I_l = \sum_h \left\lvert A_{h,l}^\top \frac{\partial L(x)}{\partial A_{h,l}} \right\rvert,
\end{equation}
where $A_{h,l}$ represents the attention matrix of the $h$-th attention head in the $l$-th layer. $x$ is the input, and $L(x)$ is the loss function. The saliency matrix $I_l$ for the $l$-th layer is obtained by averaging the values across all attention heads. Each element $I_l(i, j)$ of the matrix denotes the significance of the information flow from the $j$-th word to the $i$-th word in the prompt. 

We employ three quantitative metrics to assess the information flow: $S_{agg}$ measures the information flow of the aggregation path from previous context words to label words, $S_{dist}$ measures the information distribution from label words to the last token, and $S_{rest}$ accounts for other information flow between remaining words excluding $S_{agg}$ and $S_{dist}$. The average significance of information flow can be formulated as:
\begin{equation}
S = \frac{\sum_{(i,j) \in C} I_l(i, j)}{|C|},
\end{equation}
where $C$ is the total number of token interactions involved.\footnote{The full formulas of $S_{agg}$, $S_{dist}$, and $S_{rest}$ can be found in Appendix \ref{formula}.}

	\begin{figure}[htb]
 \centering
		\subfloat[][FPFT]{{\includegraphics[width=6cm]{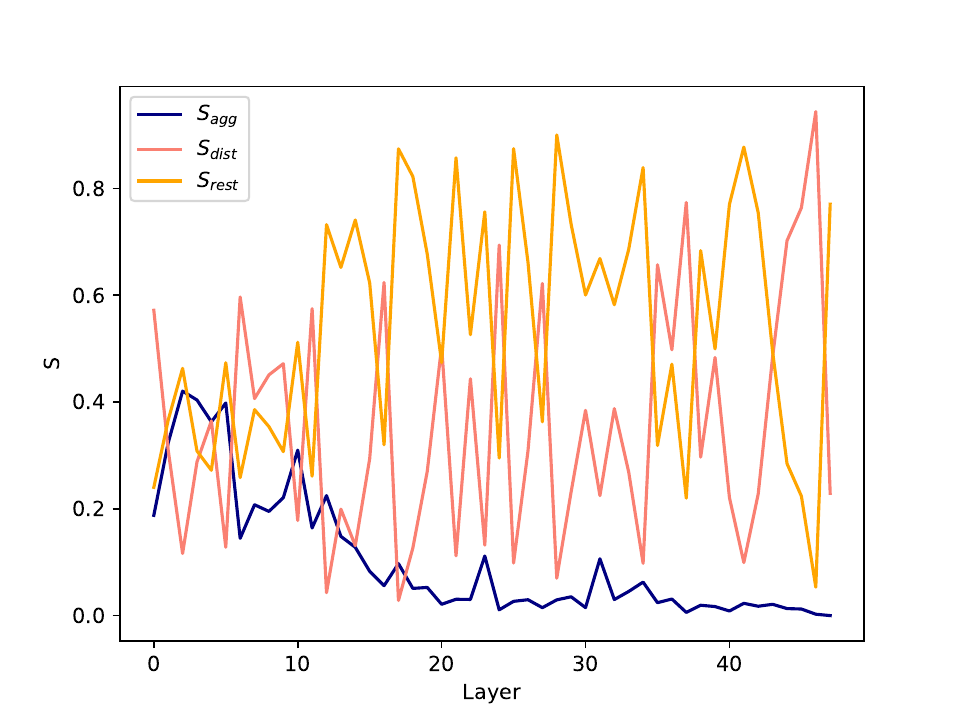} }}%
  
		\subfloat[][\method]{{\includegraphics[width=6cm]{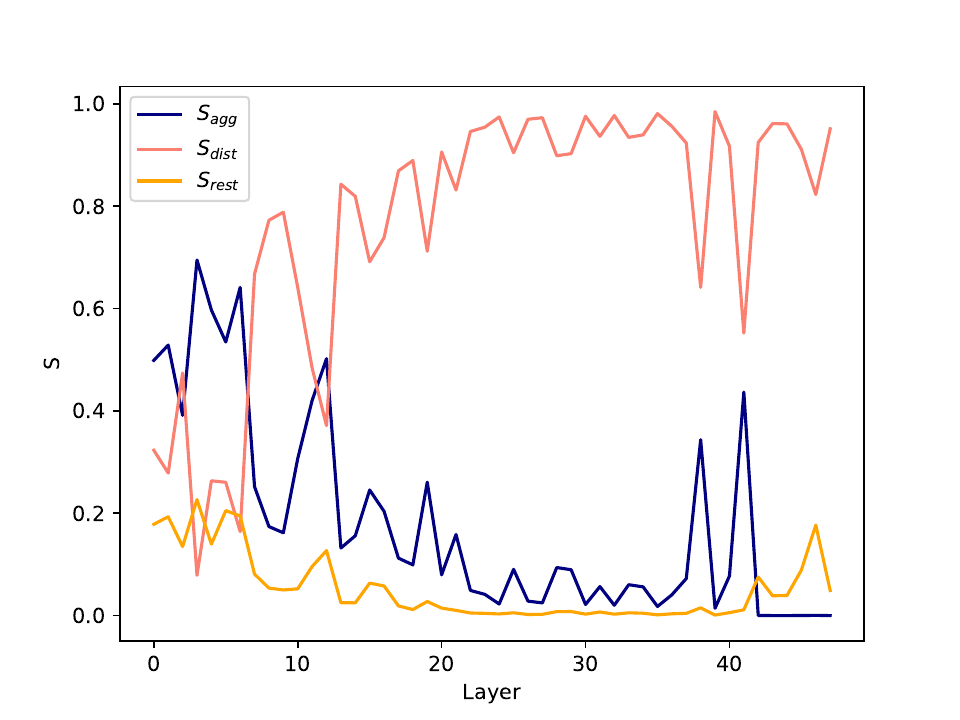} }}%
		\centering\caption{Comparison of information flow between FPFT and \method for SST-2. Both models are trained with 5 training examples per class.}\label{sali_score}
	\end{figure}
As depicted in Figure \ref{sali_score}, the information flow of \method appears more stable compared to FPFT. In FPFT, without guided navigation, tokens interact with every preceding word, leading to a trend of confusion between the information flow $S_{dist}$ and $S_{rest}$. This indicates a struggle to identify the `right' information for final prediction. Conversely, \method adheres to the information flow guided by the GNN, resulting in stable curves that depict a consistent information aggregation process, aligning with the findings of \citet{wang-etal-2023-label}. 
% The slight fluctuations around the 40th layer are attributed to the addition of the GNN layer at the 42nd layer, affecting the information aggregation paths. 
Compared to FPFT, the stable curves affirm that \method serves as a navigator, ensuring the information flows in predefined directions.

\section{Conclusion}
In this work, we propose a novel PEFT method, \method, leveraging GNN to navigate information flow within LLMs. Specifically tailored for prompt-based fine-tuning, \method significantly reduces the number of trainable parameters by simply adding a GNN layer into LLMs to guide the information flow within the prompt. \method outperforms FPFT and other PEFT methods across various classification tasks, even with few training examples. Our work offers insights into handling LLMs from a graph perspective and presents a novel application of GNNs in NLP. Future work could explore different token connectivities for GNNs or utilize GNNs to control the information flow in LLMs.

\section*{Limitation}

Although \method introduces a novel insight for NLP research, there are several limitations in our work. Firstly, \method is susceptible to the quality of the demonstrations. We find that its performance heavily relies on the selection of demonstrations when only a few training examples are available. However, this issue is alleviated with an increase in the number of training examples. Secondly, while \method builds upon the information flow of LLMs, it offers a more transparent working mechanism. However, as a black-box model, the working mechanism of the GNN layer is not investigated in this work. Thirdly, we only evaluated the performance of \method on text classification tasks, other NLP tasks are not explored in this study. We leave these limitations for future work.

\section*{Ethics Statement}
This study adhered to the ACM Code of Ethics. The datasets employed in our research are publicly accessible, and we utilized them solely for the purpose of evaluating our models. Any potential inaccuracies in the datasets are beyond our responsibility.

\section*{Acknowledgements}
We want to express our gratitude to all anonymous reviewers for
their invaluable contributions and constructive feedback. 
This work was supported by 
Center for Scalable Data Analytics and Artificial Intelligence (ScaDS.AI),
German Research Foundation (DFG) grant (SCHU 2246/14-1), Munich Center for Machine Learning (MCML), 
and China Scholarship Council (CSC).
Besides, we would like to thank Johanna Reiml for her valuable help in running additional experiments.

\bibliography{custom,anthology}
\clearpage

\appendix

\section{Hyperparameters\label{hyp}}

We present the hyperparameters for \method and other baselines in Table \ref{hyper}. The models were trained using NVIDIA A100-SXM4-40GB GPUs. Due to limited resources, the batch size was set to 1, and full parameter fine-tuning of Llama2 was implemented using 8 bits. We observed that for Llama2, \method and other PEFT methods were sensitive to the selection of prompts with very few training samples, and thus could not achieve optimal performance. To address this, we replaced these results by using another random seed to change the demonstrations in the prompt.

\begin{table*}
\centering
\begin{tabular}{l|c|c|c|c|c} 
\toprule
\textbf{Hyperparameter} & \textbf{\method} & \textbf{Prefix}          & \textbf{Adapter} & \textbf{LoRA} & \textbf{FPFT}  \\ 
\hline
learning rate          & 1e-2      & 1e-2                 & 5e-5             & 5e-4          & 5e-5           \\ 
\hline
optimizer               & Adam      & Adam                 & AdamW            & AdamW         & AdamW          \\ 
\hline
epochs                  & 50        & 50                   & 50               & 50            & 50             \\ 
\hline
early Stop              & 15        & 15                   & 15               & 15            & 15             \\ 
\hline
random seed            & \multicolumn{5}{c}{{[}0, 42, 312, 411, 412, 421, 520, 1218]}                         \\ 
\hline
virtual tokens           & -         & 40(GPT2), 150(Llama2) & \multicolumn{3}{c}{-}                            \\
\bottomrule
\end{tabular}
\caption{Hyperparameters for \method and baselines.}
\label{hyper}
\end{table*}

\section{Demonstration Templates and Label Words\label{template}}
The templates for the prompt are presented in Table \ref{temp}. $[S]$ denotes the demonstration selected to form the prompt, $[L]$ represents the label word of the demonstration, and $[S_i]$ denotes the sample to be predicted.

\begin{table*}
\centering
\scalebox{0.9}{
\begin{tabular}{l|l|l} 
\toprule
\textbf{Task} & \textbf{Template~}                                                                                                                                                              & \textbf{Label Words}                                                                                   \\ 
\midrule
SST-2         & \begin{tabular}[c]{@{}l@{}}Review:\\$[S]$\\Sentiment:\\$[L]$\\Review:\\$[S_i]$\\Sentiment:\end{tabular}              & Positive, Negative                                                                                     \\ 
\midrule
EmoC          & \begin{tabular}[c]{@{}l@{}}Dialogue:\\$[S]$\\Emotion:\\$[L]$\\Dialogue:$[S_i]$\\Emotion:\end{tabular}           & \begin{tabular}[c]{@{}l@{}}Happy, Sad, \\Angry, Others\end{tabular}                                    \\ 
\midrule
TREC          & \begin{tabular}[c]{@{}l@{}}Question:\\$[S]$\\Answer Type:\\$[L]$\\Question:\\$[S_i]$\\Answer Type:\end{tabular} & \begin{tabular}[c]{@{}l@{}}Abbreviation, Entity,\\Description, Person,\\Location, Number\end{tabular}  \\
\midrule
Amazon        & \begin{tabular}[c]{@{}l@{}}Review:\\$[S]$\\Sentiment:\\$[L]$\\Review:\\$[S_i]$\\Sentiment:\end{tabular}         & Positive, Negative                                                                                     \\ 
\midrule
AGNews        & \begin{tabular}[c]{@{}l@{}}Article:\\$[S]$\\Answer:\\$[L]$\\Article:\\$[S_i]$\\Answer:\end{tabular}             & \begin{tabular}[c]{@{}l@{}}World, Sports, \\Business, Technology\end{tabular}                          \\
\bottomrule
\end{tabular}}
\caption{Template for prompt.}
\label{temp}
\end{table*}

\section{Full Results\label{full_results_app}}

Due to space constraints, the complete results are provided in Table \ref{full_results}. Each value in the table represents the average accuracy over five experiments conducted with different random seeds.

\begin{table*}[ht]
\centering
\scalebox{0.6}{
\begin{tabular}{c|c|rcccccc|rcccccc} 
\toprule
\textbf{$k$}         & \textbf{Method} & \textbf{\#Param} & \textbf{SST-2} & \textbf{EmoC}  & \textbf{TREC}  & \textbf{Amazon} & \textbf{AGNews} & \textbf{Average} & \textbf{\textbf{\#Param}} & \textbf{SST-2} & \textbf{EmoC}  & \textbf{TREC}  & \textbf{Amazon} & \textbf{AGNews}         & \textbf{Average}  \\ 
\midrule
                     &                 & \multicolumn{7}{c|}{\textbf{GPT2-XL}}                                                                                      & \multicolumn{7}{c}{\textbf{Llama2}}                                                                                                          \\ 
\midrule
0                    & ICL             & -                & 55.44          & 6.48           & 54.68          & 53.32           & 72.12           & 48.41            & -                         & 67.55          & 9.60           & 70.36          & 94.98           & 84.14                   & 65.33             \\ 
\midrule\midrule
\multirow{7}{*}{5}   & ICL             & -                & 63.17          & 6.30           & 57.68          & 53.67           & 50.43           & 46.25            & -                         & 86.93          & 20.18          & 45.72          & 92.30           & 80.16                   & 65.06             \\ 

 \cmidrule{2-16}                    & LoRA            & 2.5M             & 91.98          & 50.60          & 75.20          & 88.80           & \textbf{85.20}  & 78.36            & 4.2M                      & \textbf{95.42} & 64.20          & \textbf{88.40} & 91.80           & 86.60                   & 85.28             \\
                     & Prefix          & 6.1M             & 59.13          & 73.46          & 32.92          & 60.00           & 75.40           & 60.18            & 39.3M                     & 50.96          & 58.56          & 21.36          & 49.36           & 25.78                   & 41.20             \\
                     & Adapter         & 15.4M            & 79.82          & 76.00          & \textbf{79.60} & \textbf{91.45}  & 81.25           & 81.62            & 198M                      & 50.92          & \textbf{84.05} & 18.80          & 49.45           & 24.80                   & 45.60             \\
                     & FPFT            & 1.6B             & 62.13          & 61.30          & 65.28          & 73.00           & 80.82           & 68.51            & 6.7B                      & 94.63          & 61.92          & 81.72          & \textbf{95.86}  & \textbf{87.58}          & 84.34             \\ 
\cmidrule{2-16}
                     & \gcn            & 2.6M             & \textbf{84.31} & 75.48          & 76.72          & 90.90           & 83.16           & \textbf{82.11}   & 16.8M                     & 94.56          & 78.30          & 83.2           & 94.00           & 86.25                   & 86.63              \\
                     & \sage           & 5.1M             & 81.95          & \textbf{78.70} & 77.92          & 88.66           & 82.88           & 82.02            & 33.6M                     & 92.91          & 80.12          & 80.80          & 95.66           & 86.06                   & \textbf{87.11}    \\ 
\midrule\midrule

 \multirow{6}{*}{10}     & LoRA            & 2.5M             & \textbf{88.08} & 53.20          & 86.40          & 90.60           & 86.80           & 81.02            & 4.2M                      & \textbf{94.73} & 63.00          & \textbf{92.80} & 92.60           & \textbf{90.40}          & 86.71             \\
 & Prefix          & 6.1M             & 51.08          & 77.58          & 38.16          & 65.94           & 61.48           & 58.85            & 39.3M                     & 50.80          & \textbf{76.98} & 21.20          & 51.42           & 26.44                   & 45.37             \\
                     & Adapter         & 15.4M            & 86.70          & 70.65          & \textbf{87.40} & 90.60           & 86.15           & 84.30            & 198M                      & 50.92          & 85.60          & 41.00          & 52.20           & 52.15                   & 56.37             \\
                     & FPFT            & 1.6B             & 69.01          & 71.90          & 52.48          & 75.82           & 81.34           & 70.11            & 6.7B                      & 92.91          & 68.06          & 84.24          & 96.22           & 88.64                   & 86.01             \\ 
\cmidrule{2-16}
                     & \gcn            & 2.6M             & 84.63          & \textbf{83.97} & 74.80          & 91.57           & \textbf{87.00}  & \textbf{84.39}   & 16.8M                     & 91.86          & 70.75          & 82.40          & \textbf{96.35}  & 89.30                   & 84.99             \\
                     & \sage           & 5.1M             & 87.41          & 77.98          & 78.28          & \textbf{91.90}  & 84.52           & 84.02            & 33.6M                     & 94.06          & 76.02          & 83.96          & 95.76           & 87.64                   & \textbf{87.49}    \\ 
\midrule\midrule
   \multirow{6}{*}{20}    & LoRA            & 2.5M             & 85.09          & 69.00          & 86.00          & \textbf{94.00}  & \textbf{89.20}  & 84.66            & 4.2M                      & 95.64          & 70.80          & 83.60          & \textbf{96.20}  & \textbf{\textbf{90.60}} & 87.37             \\
 & Prefix          & 6.1M             & 56.68          & \textbf{83.28} & 39.20          & 61.22           & 80.62           & 64.20            & 39.3M                     & 50.57          & 78.70          & 27.92          & 52.08           & 26.30                   & 47.11             \\
                     & Adapter         & 15.4M            & 88.42          & 74.65          & \textbf{89.00} & 89.45           & 86.50           & 85.60            & 198M                      & 50.92          & \textbf{85.80} & 18.80          & 56.40           & 24.80                   & 47.34             \\
                     & FPFT            & 1.6B             & 73.10          & 70.72          & 68.36          & 77.40           & 80.44           & 74.00            & 6.7B                      & \textbf{95.32} & 69.96          & \textbf{88.08} & 95.52           & 89.04                   & 87.58             \\ 
\cmidrule{2-16}
                     & \gcn            & 2.6M             & 86.93          & 76.23          & 79.67          & 92.70           & 86.07           & 84.32            & 16.8M                     & 94.78          & 75.25          & 84.80          & 96.00           & 89.30                   & 88.27             \\
                     & \sage           & 5.1M             & \textbf{88.67} & 78.96          & 82.52          & 92.02           & 86.24           & \textbf{85.68}   & 33.6M                     & 94.56          & 79.92          & 84.56          & 95.64           & 88.54                   & \textbf{88.64}    \\ 
\midrule\midrule
    \multirow{6}{*}{50}          & LoRA            & 2.5M             & 89.45          & 74.80          & 54.80          & \textbf{93.60}  & \textbf{91.80}  & 80.89            & 4.2M                      & 93.12          & 72.40          & \textbf{94.40} & 95.40           & \textbf{91.60}          & 89.20             \\
  & Prefix          & 6.1M             & 50.90          & \textbf{79.78} & 26.72          & 74.42           & 74.40           & 61.24            & 39.3M                     & 50.48          & 76.22          & 28.08          & 50.96           & 27.60                   & 46.67             \\
                     & Adapter         & 15.4M            & 86.75          & 77.85          & \textbf{91.60} & 90.50           & 88.75           & 87.09            & 198M                      & 50.92          & 76.80          & 44.40          & 49.45           & 33.45                   & 51.00             \\
                     & FPFT            & 1.6B             & 70.60          & 71.68          & 76.40          & 67.84           & 83.10           & 73.92            & 6.7B                      & \textbf{95.46} & 74.20          & 91.92          & 95.82           & 90.48                   & 89.58             \\ 
\cmidrule{2-16}
                     & \gcn           & 2.6M             & 89.49          & 79.50          & 87.93          & 92.40           & 87.43           & \textbf{87.35}   & 16.8M                     & 95.07          & \textbf{83.05} & 88.70          & 95.85           & 90.80                   & \textbf{90.81}    \\
                     & \sage           & 5.1M             & \textbf{90.14} & 75.70          & 87.96          & 93.26           & 87.30           & 86.87            & 33.6M                     & 94.72          & 79.04          & 90.72          & \textbf{96.00}  & 90.68                   & 90.23             \\ 
\midrule\midrule
 \multirow{6}{*}{100}& LoRA            & 2.5M             & 89.22          & \textbf{84.00} & 88.40          & 93.20           & 84.80           & 87.92            & 4.2M                      & 92.66          & \textbf{86.60} & 94.80          & 95.40           & 67.60                   & 87.41             \\
 & Prefix          & 6.1M             & 56.26          & 72.28          & 32.04          & 69.48           & 51.18           & 56.25            & 39.3M                     & 49.11          & 76.20          & 40.28          & 52.38           & 26.82                   & 48.96             \\
                     & Adapter         & 15.4M            & 86.93          & 82.85          & \textbf{92.00} & 92.40           & 87.60           & 88.36            & 198M                      & 58.83          & 84.95          & 84.00          & 68.10           & 24.80                   & 64.14             \\
                     & FPFT            & 1.6B             & 72.82          & 73.42          & 68.56          & 78.74           & 84.86           & 75.68            & 6.7B                      & \textbf{95.07} & 76.06          & \textbf{96.20} & \textbf{96.20}  & \textbf{91.04}          & \textbf{90.91}    \\ 
\cmidrule{2-16}
                     & \gcn            & 2.6M             & 89.41          & 81.30          & 90.20          & 92.67           & 87.97           & 88.31            & 16.8M                     & 94.27          & 81.20          & 91.60          & 96.00           & 90.80                   & 90.77             \\
                     & \sage         & 5.1M             & \textbf{90.46} & 80.16          & 91.12          & \textbf{93.28}  & \textbf{88.58}  & \textbf{88.72}   & 33.6M                     & 94.45          & 81.20          & 90.88          & 96.08           & 90.78                   & 90.68             \\ 
\midrule\midrule
 \multirow{6}{*}{200}     & LoRA            & 2.5M             & \textbf{90.83} & 80.80          & 90.80          & 82.00           & 86.20           & 86.13            & 4.2M                      & 91.29          & \textbf{86.80} & 93.60          & 95.80           & 90.40                   & 91.32             \\
 & Prefix          & 6.1M             & 50.92          & 80.18          & 69.80          & 59.80           & 79.08           & 67.96            & 39.3M                     & 48.35          & 81.72          & 45.68          & 52.28           & 27.54                   & 51.11             \\
                     & Adapter         & 15.4M            & 88.65          & 80.70          & \textbf{96.60} & 92.30           & \textbf{89.80}  & 89.61            & 198M                      & 50.92          & 85.05          & 88.20          & 49.45           & 81.50                   & 67.57             \\
                     & FPFT            & 1.6B             & 68.97          & 73.70          & 80.16          & 74.82           & 85.34           & 76.60            & 6.7B                      & \textbf{95.64} & 79.90          & \textbf{96.76} & 96.12           & \textbf{91.44}          & 91.97             \\ 
\cmidrule{2-16}
                     & \gcn            & 2.6M             & 90.67          & 78.82          & 91.88          & 92.94           & 89.20           & 88.70            & 16.8M                     & 95.36          & 82.85          & 95.50          & \textbf{96.45}  & 91.05                   & \textbf{92.24}    \\
                     & \sage          & 5.1M             & 90.46          & \textbf{82.68} & 92.32          & \textbf{93.44}  & 89.28           & \textbf{89.64}   & 33.6M                     & 95.30          & 81.94          & 94.76          & 95.96           & 90.68                   & 91.73             \\
\bottomrule
\end{tabular}}
\caption{Results with different training methods (accuracy). $k$ denotes the number of training examples per class. \#Param denotes the number of trainable parameters. The best scores under the same circumstances of training examples are highlighted with \textbf{bold}.}
\label{full_results}
\end{table*}

\section{Formula of Saliency Score\label{formula}}
We utilize $l$ to denote the label words, such as `Positive' and `Negative', while $f$ represents the final token, such as `:'. Additionally, $t$ denotes other tokens excluding label and final tokens.

$S_{agg}$ calculates the mean significance of information flow from the previous context words to label words:
\begin{equation}
\begin{split}
S_{agg} = \frac{\sum_{(i,j) \in C_{tl}} I_l(i, j)}{|C_{tl}|},\\
C_{tl} = \left\{(l_k, j) : k \in [1, C], j < l_k\right\}.
\end{split}
\end{equation}

$S_{dist}$ calculates the mean significance of information flow from the label words to the final token:
\begin{equation}
\begin{split}
S_{dist} = \frac{\sum_{(i,j) \in C_{lf}} I_l(i, j)}{|C_{lf}|},\\
C_{lf} = \left\{(f, l_k) : k \in [1, C]\right\}.
\end{split}
\end{equation}

$S_{rest}$ calculates the mean significance of information flow among the rest words, excluding $S_{agg}$ and $S_{dist}$:
\begin{equation}
\begin{split}
S_{rest} = \frac{\sum_{(i,j) \in C_{tt}} I_l(i, j)}{|C_{tt}|},\\
C_{tt} = \left\{(i, j) : j < i\right\}-C_{tl}-C_{lf}.
\end{split}
\end{equation}

\end{document}